\documentclass[10pt, final, journal, a4paper, oneside, twocolumn]{IEEEtran}
\IEEEoverridecommandlockouts
\usepackage{cite}
\usepackage{amsmath,amssymb,amsfonts}
\usepackage{algorithmic}
\usepackage{algorithm} % used for the formation of the pseudo code
\usepackage{graphicx}
\usepackage{textcomp}
\usepackage{xcolor}
\usepackage[unicode, hidelinks]{hyperref} % For hyperlinks in the generated document.
\usepackage{multirow} % For table editing
\usepackage{stfloats} % To be able to place table* on the bottom
\usepackage{arydshln} % offers you the \hdashline and \cdashline commands which are the dashed counterparts of \hline and \cline

\ifCLASSOPTIONcompsoc
\usepackage[caption=false,font=normalsize,labelfon
t=sf,textfont=sf]{subfig}
\else
\usepackage[caption=false,font=footnotesize]{subfi
g}
\fi

\def\BibTeX{{\rm B\kern-.05em{\sc i\kern-.025em b}\kern-.08em
    T\kern-.1667em\lower.7ex\hbox{E}\kern-.125emX}}
    
\begin{document}

\title{Reinforcement Learning for Solving Robotic Reaching Tasks in the Neurorobotics Platform\\
}

\author{\IEEEauthorblockN{M{\'a}rton Sz{\'e}p,}
\IEEEauthorblockN{Leander Lauenburg,}
\IEEEauthorblockN{Kevin Farkas,}
\IEEEauthorblockN{Xiyan Su,}
\IEEEauthorblockN{Chuanlong Zang} \\
\IEEEauthorblockA{Technical University of Munich, Germany \\
\texttt{\{marton.szep, leander.lauenburg, kevin.farkas, tim.su, chuanlong.zang\}@tum.de}}
}

\maketitle

\begin{abstract}
In recent years, reinforcement learning (RL) has shown great potential for solving tasks in well-defined environments like games or robotics. This paper aims to solve the robotic reaching task in a simulation run on the Neurorobotics Platform (NRP). The target position is initialized randomly and the robot has 6 degrees of freedom. We compare the performance of various state-of-the-art model-free algorithms. At first, the agent is trained on ground truth data from the simulation to reach the target position in only one continuous movement. Later the complexity of the task is increased by using image data as input from the simulation environment. Experimental results show that training efficiency and results can be improved with appropriate dynamic training schedule function for curriculum learning.
\end{abstract}


\section{Introduction}
Reinforcement learning (RL) is a paradigm in the field of machine learning that has recently gained tremendous interest \cite{sutton2018reinforcement} and achieved notable success in robotic control policy learning tasks \cite{gu2017deep, tai2017virtual, lan2021learning}. However, real-world applications are still scarce and challenging. The primary challenges, which prevent these methods from being widely adopted in the industry, are the safety issues in the exploration phase and the overall low training efficiency. RL algorithms adapt their policy based on the experience gathered from informative rewards. Unfortunately, random exploration of the action space usually yields sparse rewards. The reason is that the probability of finding a beneficial action to reach a specific goal is really low in high dimensional state and/or action spaces. Accordingly, an agent in such a setting requires longer training times to combat the odds.

Skinner \cite{skinner1958reinforcement} introduced the concept of successive approximation based on training a pigeon to bowl. Instead of rewarding the target behavior directly, he found that positively reinforcing any behavior that is approximately in the direction of the desired final goal (i.e. appropriate guidance) has the effect of facilitating the learning process of the pigeon. This concept has resurfaced under the name of curriculum learning \cite{bengio2009curriculum} with the main idea that the skills learned from basic preliminary tasks can help an agent perform later better in more complex tasks.

The main contributions of our study are as follows: First, we set up a self-contained simulation-based RL environment in the Neurorobotics Platform\footnote{\url{https://neurorobotics.net/}} (NRP) for rapid prototyping and scalable RL experiments. Secondly, we train multiple state-of-the-art model-free agents to solve the reaching task based on ground truth data. We start with a simplified version of the task and make it iteratively more difficult to guide the agent. Lastly, the pipeline is extended to take image data as input.

\section{Related Work}
The main goal of this study is to solve the reaching task in the NRP with RL while leveraging  curriculum learning. Although an extensive survey of other approaches for RL in robotic control is beyond the scope of this study, we will review a few recent efforts.

Various research groups have made substantial progress towards the development of continuous RL for robot control. The long training times in RL are a testament to how difficult it is to train agents to behave as expected with pure trial-and-error mechanisms \cite{plappert2018multi, andrychowicz2017hindsight}. One conventional method to accelerate the learning is to combine RL with Imitation Learning \cite{nair2018overcoming}. This is done by creating demonstration state-action pairs and using them either for supervised learning (behavior cloning) or incorporating them directly to the policy improvement step of the RL algorithm.

Another popular method related to curriculum learning is hindsight experience replay (HER) \cite{andrychowicz2017hindsight}. HER can be considered implicit curriculum learning as it allows for skipping complicated (manual) reward engineering. It improves learning efficiency by reinterpreting unsuccessful experiences in achieving a particular goal into successful ones by adapting the goal to the result of the chosen action. Nair et al. \cite{nair2018overcoming} report that the use of HER provides an order of magnitude of speedup for simulated robotics tasks. In this study we investigate this in combination with state-of-the-art model-free off-policy algorithms.

Kerzel et al. \cite{kerzel2018accelerating} accelerate the learning by dynamically adjusting the target size. The dynamical adjustment ensures that the robot gets some positive rewards also at the beginning of the training, when it has little knowledge of the task. Our approach also incorporates this by automatically decreasing the required distance threshold to the target based on performance. We also investigate the possibility of gradually increasing the number of actuated joints throughout the training. Luo et al. \cite{luo2020accelerating} propose a method of continuous curriculum learning that aims to rectify an issue of the previous approach (defining discrete sub-tasks) residing in the possible unstable performance of the agent, as it needs to adjust its skills to a new, stricter task.

\section{Background and Methodology}
Our work builds upon model-free RL algorithms. Hence, we first give a brief introduction to the background of RL, the algorithms we used and compared, and, finally, we provide a short description of the reaching task.

\subsection{Reinforcement Learning}
The RL-based reaching problem satisfies the Markov property, thus it can be formalized as a Markov Decision Process (MDP) \cite{sutton2018reinforcement}. The tuple $(S, A, P, R, \gamma)$ models the process, where $S$ and $A$ denote the finite state and action space, respectively. Along these lines we can talk about a finite MDP. $P$ is the state transition probability matrix $$P: p(s' | s, =a) = Pr\{s_{t+1}=s' | s_t=s, a_t=a\}$$ which represents the probability to reach a state $s'$ given the current state $s$ by taking the action $a$. $R$ is the reward space for the expected rewards $r(s, a) = \mathbb{E}[r_{t+1} | s_t=s, a_t=a]$ associated with state-action pairs and $\gamma \in [0, 1]$ is the discount factor. The agent aims to learn a deterministic policy $\pi(a_t | s_t)$ specifying the next action based on the current state. The learning process is taking place by interacting with the environment with the goal to maximize the future $\gamma$-discounted expected return $R_t = \mathbb{E}[\sum_{i=0}^\infty \gamma^i r_{t+i+1}]$ at every time step $t$.

\subsection{Algorithms} \label{sec:algorithms}
One important distinction of RL algorithms is whether the agent has (or learns) some kind of environment model. Often it is challenging to build accurate models of robotic tasks, so we restrict our interest to model-free algorithms. Generally, model-free algorithms adopt two different approaches to represent and train an agent. The first one, \textit{policy optimization}, aims to find the parameters $\theta$ of the explicit policy representation $\pi_\theta(a | s)$ by minimizing an objective function. The second approach, \textit{Q-Learning}, indirectly optimizes for agent performance by training an approximator $Q_\theta(s, a)$ for the optimal action-value function. While policy optimization is in general more stable and reliable, Q-learning is more sample efficient. Fortunately, they are not incompatible, and hybrid models interpolating between policy optimization and Q-learning have achieved outstanding performance in continuous state and action space. This motivated us to use them in our experiments, which is why we will briefly present them in the followings.

\begin{algorithm}
\caption{Reaching Logic}
\label{alg:NextEpoch}
\begin{algorithmic} 
\REQUIRE $\textit{thresh} \geq 0$
\REQUIRE $\textit{max } \text{tries} \geq 0$
\FOR{\textit{epsiodes}}
\STATE $\textit{try } = 0$
\STATE $ \textit{obs} \leftarrow \textit{setup()}$
\WHILE{$\textit{\textbf{not}  done}$}
\STATE$ \textit{act} \leftarrow \text{agent choose } \textit{action(obs)}$
\STATE$ \textit{obs} \leftarrow \text{client } \textit{execute(action)}$
\STATE$ \textit{rew} \leftarrow \text{reward(obs)}$
\IF{$\textit{rew} < \textit{thresh} \textbf{ or } \textit{try} \geq \textit{max} $}
\STATE $\textit{done} = \text{True }$
\ELSE
\STATE $\textit{try } += 1$
\ENDIF
\STATE $\text{agent } \textit{remember(obs, act, rew, done)}$

\STATE $\text{agent } \textit{learn()}$

\ENDWHILE

\STATE $\textit{thresh} \leftarrow  \text{threshold\_scheduler}(\textit{thresh},  \textit{reward\_history})$

\ENDFOR
\end{algorithmic}
\end{algorithm}
Generally speaking the robot reaching problem follows the logic depicted in Algorithm~\ref{alg:NextEpoch}. Our algorithms are all based on the actor-critic architecture. The critic network takes a state and an action as input and tells the actor-network if the chosen action was "good" or "bad". The actor-network handles how the agent selects an action.

While the implementation of the learning function and the action selection distinguishes the chosen RL algorithms, all of them rely on a replay buffer. The replay buffer functions as the agent's memory. At the start, the replay buffer is filled up with the initial state, the action, the reward, the final state, and a done flag for each episode. Later the model replays and learns from these saved transitions.

When the agent interacts with the environment, the transition data sequences are highly correlated in time. This could cause overfitting and bad convergence behavior if directly used for training. Using a replay buffer offers remedy to this issue. The agent can randomly sample mini-batch data from the experience replay buffer during training, so that the sampled data can be considered unrelated.

\subsubsection{Deep Deterministic Policy Gradient (DDPG)}
DDPG is an enhanced version of Deterministic Policy Gradient (DPG) \cite{silver2014deterministic}, which originates from Policy Gradient (PG) \cite{sutton2000policy}.

DPG chooses a deterministic policy to avoid frequent sampling actions in the high-dimensional action space as the original PG does in each algorithm step. The model's behavior is directly obtained through the policy function $\mu$ to determine the action. This function $\mu$ is the optimal behavior policy, which is no longer a random policy $\pi$ in PG \cite{silver2014deterministic}.

DDPG merges deep learning techniques into DPG. DPG realizes the policy function $\mu$ and the Q-function via linear regression. DDPG, in turn, implements those two functions via neural networks, namely the policy network and the Q-network. In DDPG, the way to implement and train Q-function is adopted from DQN (Deep Q-Network) \cite{lillicrap2015continuous, mnih2013playing}. Accordingly, DDPG's critic network is based on a Q-Network and target Q-network, while its actor network uses a policy network and a target policy network.

\subsubsection{Twin Delayed DDPG (TD3)}
TD3 is an extension of DDPG addressing the issue of the Q-function dramatically overestimating the Q-values causing policy breaking. Fujimoto et al. \cite{fujimoto2018addressing} introduce three tricks to counter this phenomenon. The first one is target policy smoothing which adds clipped noise to each action dimension, serves as a regularizer for the algorithm, and avoids incorrect sharp peaks of the Q-function. Second, clipped double-Q learning approximates two Q-functions and always uses the small Q-value as regression target. This helps to fend off overestimation in the Q-function. Third, delayed policy updates the policy and target networks less frequently than the Q-functions are updated. This helps mitigate the policy fluctuations that commonly arise because a policy update changes the target, which acts back on the policy through the Q-functions.

\subsubsection{Soft Actor Critic (SAC)}
The most important contribution of SAC is its modified RL objective function. Usually, the RL algorithms aim only to maximize the lifetime reward. SAC additionally tries to maximize the entropy of the policy to promote exploration \cite{Haarnoja}.

To this end, SAC extends the general actor-critic architecture with a value-network. The value network evaluates a state independently from the taken action. Therefore, the network has the state as single input. Additionally, in contrast to the DDPG and TD3, the actor-network outputs a mean and standard deviation \cite{Haarnoja}. Wherefore, the actor-network has two heads. One head calculates the mean, and the other calculates the standard deviation. The standard deviation is then clamped to limit the spread of the distribution. The network then samples from the normal distribution and returns it as its output. 

Given an observation, the SAC agent chooses an action using the action network. After executing an action, the agent fills the replay buffer with the initial observation, the chosen action, the calculated return, and the final observation. The agent starts the learning process after the replay buffer reaches a predefined number of entries. In the learning step, the agent calculates the value of the given state, updates the policy using the critic network, and updates the parameters of the three networks accordingly.

\subsection{Reaching Task} \label{sec:reaching_task}
In the field of robotic manipulation, the reaching task is a well-defined problem that aims to find a valid motion trajectory to put a part of the robot (typically the end effector) to a desired goal position within its workspace. Taking a look at recent publication to the topic of end effector reaching task, a major part of them assigns the orientation of the end effector a subordinate role \cite{pham2018optlayer, aumjaud2020reinforcement, aumjaud2021rl_reach}. Experiments considering both, position and orientation of end effectors, typically try to solve a kind of grasping, pushing or other related tasks \cite{levine2018learning, luo2020accelerating, gu2017deep}.

Just as in other fields of RL, minimizing the distance from end effector to target position is achieved by maximizing the outcome of a given reward function. The definition of a reward function can be done in different ways. A common subdivision of reward functions is sparse and dense. Sparse reward functions only give a positive reward on reaching the target position. Dense reward functions already reward movements in the right direction. A target is considered to have been reached when the distance between end effector and goal position falls below a certain threshold. The challenge with a dense reward function lies in the modelling, since various factors influence the learning behaviour of the agent. The sparse reward function, on the other hand, is easy to define due to its binary structure. In our experiments, both the sparse reward function and a dense reward function are investigated. The used reward functions are given in Tab.~\ref{tab:reward_functions}.

\begin{table}[tbp]
\caption{Definitions of used reward functions. The rewards $r$ are calculated depending on the ratio of distance $d$ to threshold $\tau$}
\begin{center}
\begin{tabular}{|c|c|}
\hline
\textbf{Reward Function}  &\textbf{Definition}\\
\hline
sparse reward & 
$r= 
\begin{cases} 
1 & \mbox{if } d < \tau \\
0 & \mbox{else }
\end{cases}$ \\
\hline
dense reward &
$r= 
\begin{cases} 
1 & \mbox{if } d < \tau \\
-d & \mbox{else }
\end{cases}$ \\
\hline
%extra dense reward &
%$r= 
%\begin{cases} 
%3 & \mbox{if } d < \frac{\tau}{2} \\
%1 & \mbox{if } d < \tau \\
%-d & \mbox{else }
%\end{cases}$ \\
%\hline
\end{tabular}
\label{tab:reward_functions}
\end{center}
\end{table}

Furthermore, different approaches can be distinguished based on how many discrete movement steps the agent is allowed to do to reach the goal. Often, the number of steps needed to successfully complete an episode is also used to evaluate and compare different models \cite{luo2020accelerating, aumjaud2020reinforcement}. However, since in the case of several available movement steps it tends to be more difficult for RL models to determine which steps were advantageous and which not, a discount factor is usually introduced to allow later steps to be included to a greater extent in the evaluation of the actions \cite{sutton2018reinforcement}. To reduce the number of hyperparameters we decided to approach the goal in only one step per episode. In addition, this simplifies the creation of dense reward functions.

\begin{figure}[!b]
\centering
\subfloat[Initial pose]{\includegraphics[width=0.225\textwidth]{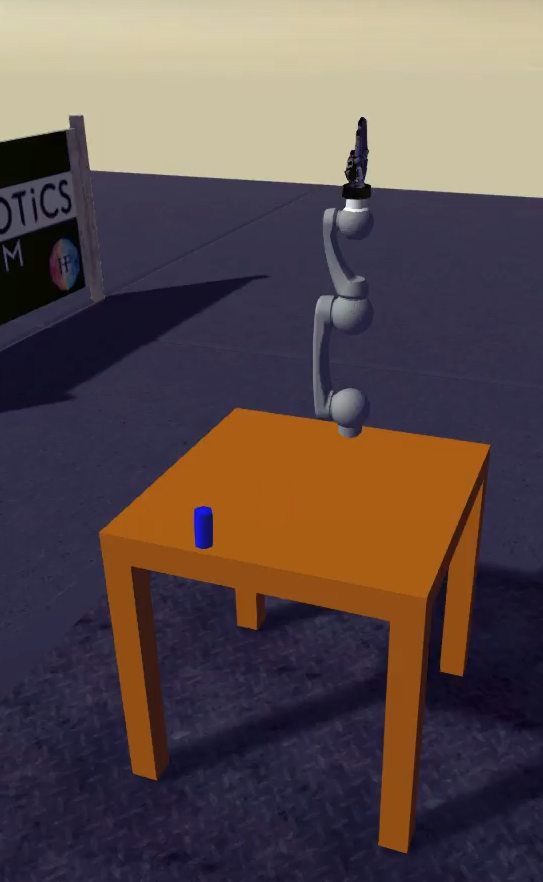}
\label{fig:reaching1}}
\hfil
\subfloat[Reaching step]{\includegraphics[width=0.225\textwidth]{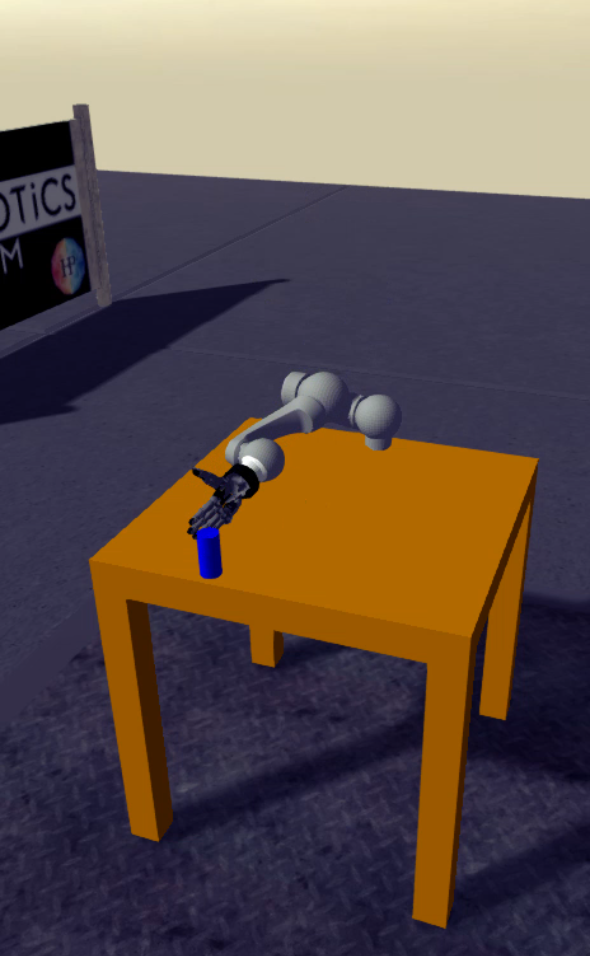}
\label{fig:reaching2}}
\caption{Reaching episode using the HoLLiE robot arm. The blue cylinder represents the target. The position of the cylinder on the table is initialised randomly in each episode. (a) The initial pose of the arm. (b) Robot posture after the agent sets the joint angles such that the distance of the positions between the cylinder and the end effector is below a required threshold. In this case, the episode is considered to be successful.}
\label{fig:reaching}
\end{figure}
In our setup, the initial position for all experiments is defined as the zero position of the 6 DoF HoLLiE robot arm (Fig.~\ref{fig:reaching1}). The reaching task is successfully completed when the center of the hand is close enough to the blue cylinder placed at a random position on the table (Fig.~\ref{fig:reaching2}).
The observation space for all investigated models has the same definition. It includes the end effector position (3 dimensions), cylinder position (3 dimensions), and all joint states (6 dimensions), resulting in a 12-dimensional vector for the entire observation space.

\section{Experiments}
In this section, we first introduce the environmental setup. The used robot model is the standard 6 degrees of freedom (DoF) HoLLiE robot arm model \cite{hermann2013hardware} operating with Gazebo as physics simulator under the Robot Operating System (ROS) framework. We set up a Python interface between the RL algorithms and the agent environment to control and observe the robot model. To reduce latency, and because the environmental components and the RL algorithms are encapsulated in Docker containers, we additionally establish a communication channel via Google Remote Procedure Call (gRPC) \cite{Marculescu2015} to the provided interface.
We then describe the performance and efficiency of the three investigated model-free RL algorithms which are learning from ground truth data. Furthermore, we illustrate an expansion of the algorithmic setup, which can learn from images captured by a top-view camera as input.  

\subsection{Environmental Setup}
The simulation is based on the 6 DoF HoLLiE robot arm model included in the NRP. The primary purpose of the NRP is to connect artificial brain models with simulated embodiments in a dynamic Gazebo environment \cite{NrpInitialReport2017}. However, in this study, we do not focus on artificial brain models and therefore only use the well-prepared Gazebo environment. 

We use a dockerized version of the NRP that is split into two containers, the frontend and the backend. The frontend container provides various instruments for taking full control of experiments in terms of initial setup, adjustments, tracking and visualization over a browser-based interface. The backend container holds all core components including model, physics engine and a control interfaces. The control interface coincidently acts as API from the Python to ROS and vice-versa. Although the backend container of the NRP framework provides all necessary components for running experiments, the frontend is used for visual observation and documentation purposes.

\begin{figure}[tbp]
\centerline{\includegraphics[width=\linewidth, trim=2.5cm 8.5cm 1.8cm 2.3cm]{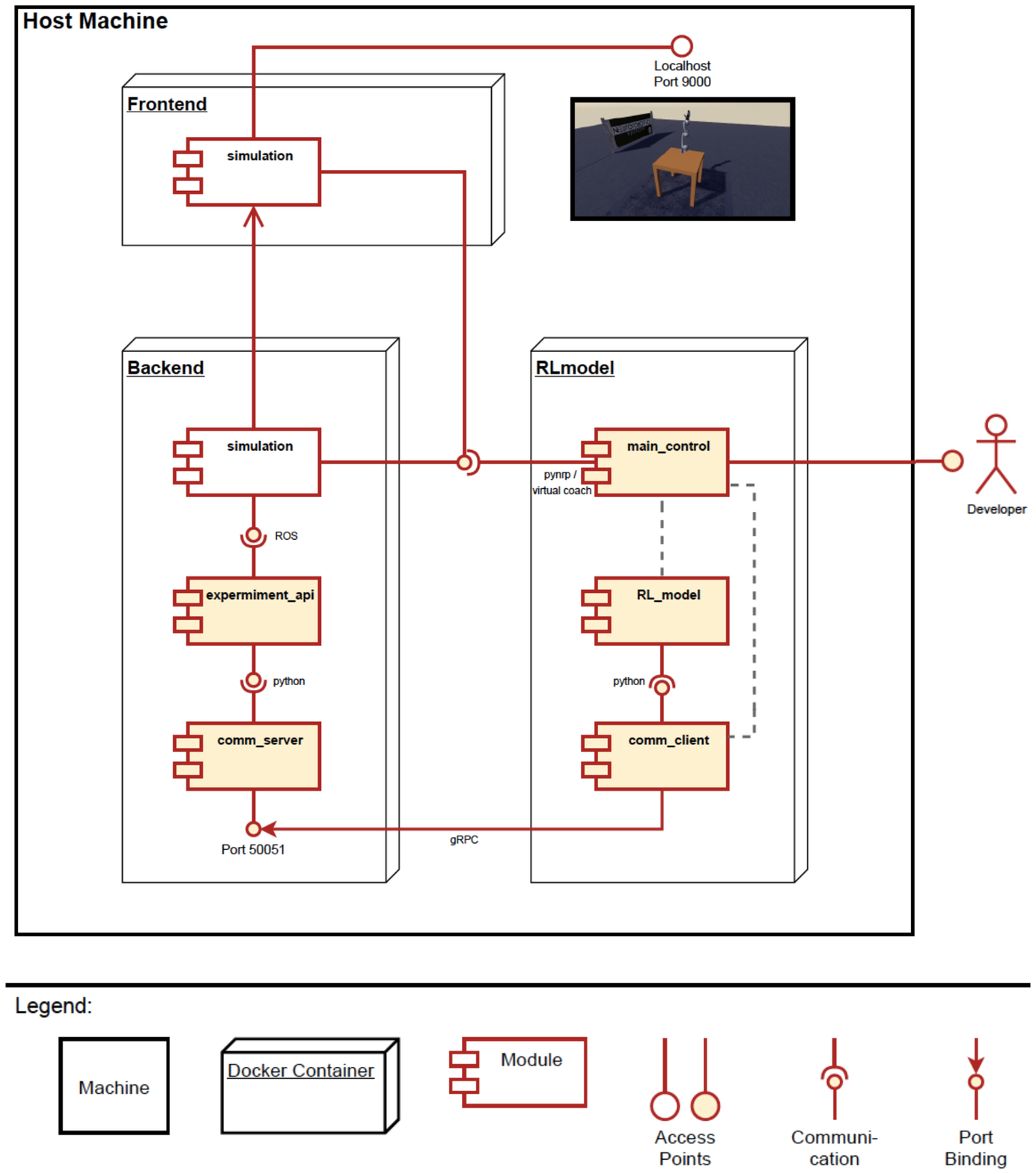}} % left bottom right top
\caption{Arrangement of the core components of the designed setup. The top and left 3d-boxes are representing the frontend and backend of the dockerized NRP framework. The right 3d-box represents the docker container containing RL algorithms. Both containers are connected via a communication channel using gRPC. Control signals to the robot are carried to the underlying simulation via a python interface (\texttt{experiment\_api}) which decodes them to ROS commands.}
\label{fig:software_stack}
\end{figure}

The investigated RL algorithms are encapsulated in a separate Docker container. For the communication between the RL models and the robot simulation we use gRPC \cite{Marculescu2015}. Since the robot environment provides the service to execute certain actions on the robot model we decide to place the gRPC server inside the backend container. The RL algorithm acts as a client which requests to do actions and observations on the robot model. The whole setup with all its building blocks is represented in Fig.~\ref{fig:software_stack}.

\subsection{Learning from ground truth data}

When learning from ground truth data, we directly feed the cylinder location to the agent. Following the logic in Algorithm~\ref{alg:NextEpoch}, the setup function resets the robot to the initial pose and returns randomly chosen cylinder coordinates on the table. Accordingly, the agent's task is find an approximation for the inverse kinematics -- which set of joint angles correspond to a given end effector location in the cartesian space. 

\subsubsection{Custom implementations} \label{sec:custom}

At first, we implemented SAC, DDPG, and TD3 from scratch. Therefore, we programmed the algorithm-specific networks, memory buffers, agent learning procedures, and custom environments. All implementations of the three algorithms were highly successful when applied the classical cart pole game as a sanity check. In the cart pole game, the agent learns to balance a stick by moving the cart to the left or the right. However, when applying the algorithms to the actual reaching problem, the performance varied considerably.

When applying the SAC to the reaching task, the agent constantly collapsed into a small subset of end positions. Although those positions maximized the average reward -- the robot moved the end effector towards the center and close to the table surface -- SAC performed particularly poorly. Considering that one of the key aspects of the SAC algorithm is its reduced number of hyperparameters, this was especially surprising. However, extensive reviews, significantly increasing the exploration factor, general tuning, observation- and action space variation did not show significant performance increase.

Our custom DDPG algorithm performed considerably better. However, we admittedly had to use some tricks to achieve this. First, we extended the algorithm to use HER. Secondly, we applied clipping to the output of the action network to keep all executed actions within the defined action space. Third, we added a small probability of 10\% to the agent to act randomly to balance exploration and exploitation.

\begin{figure}[tp]
    \centering
    \includegraphics[width=0.9\linewidth]{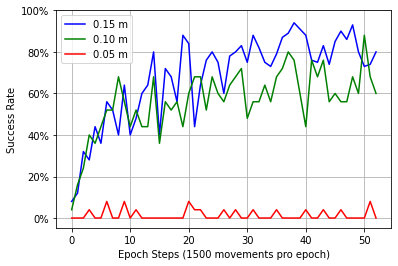}
    \caption{Three DDPG models were trained and evaluated for thresholds of 0.15~m, 0.10~m, and 0.05~m. A huge drop in performance can be observed, comparing the models trained for 0.10~m and 0.05~m. This behavior is particularly present in DDPG, that’s why its extension TD3 tries to tackle that issue.}
    \label{fig:output_ddpg}
\end{figure}

The results in Fig.~\ref{fig:output_ddpg} show that our custom DDPG algorithm is susceptible to the chosen threshold. For large thresholds, the agent converges and shows quite good results. However, for relatively small thresholds, the agent seems to face severe difficulties and can hardly learn anything.

From an algorithm's perspective, TD3 is a direct successor of DDPG with considerable improvements. Accordingly, TD3 should always perform not worse than DDPG.\cite{fujimoto2018addressing} Therefore, the DDPG algorithm is no longer implemented in the later sections to keep a more efficient workflow. But all in all, custom DDPG shows meaningful results when applied to the reaching task.

The custom implementation of the TD3 algorithm is based on \cite{fujimoto2018addressing} and does not exploit HER. As described in section~\ref{sec:algorithms}, the TD3 agent comprises an actor and critic network trained in parallel. To facilitate the training process and reduce vanishing/exploding gradient problems, we apply the weight initialization method described in \cite{he2015delving}. The environment uses the state, action space definitions, and dense reward function described in section~\ref{sec:reaching_task}. The training process of this agent has intermittent policy evaluation steps, where the action noise (used for exploration) is set to zero. Without further guidance (no action space restrictions), the reaching task proves to be too difficult to the agent. It learns a policy by averaging the actions in the replay buffer: the end effector is always placed in the middle of the table plane. To help the agent become acquainted with the task, we constrain the action space and decrease the number of actuated joints to three. The low and high boundaries of the constrained action space for using only three joints are shown in \eqref{eq: action_3joints}. The idea is that multiple solutions exist for the inverse kinematics, and we are only interested in finding one. Hence it makes sense to reduce the dimensionality of the task. This guidance is seemingly simple, but it incorporates expert knowledge from understanding the task, hence requires task-specific manual work. By all means, this primitive form of curriculum learning already has a significant impact on the agent's performance.

In addition, the intermittent evaluation steps during the training process of the custom TD3 agent allow for the computation of the success rate with an adjustable evaluation frequency. To decrease training time and increase the overall model performance, we introduced threshold scheduling as a form of curriculum learning. We systematically increase task complexity by starting with a large threshold (maximum allowed distance to the target) and decreasing it every time the agent reaches a success rate of 95\%. The intermittent evaluation steps comprised 100 episodes and were performed every 1000 steps during the training process. Tab.~\ref{tab:td3_custom} shows the results of a TD3 agent trained with the method described above for 25000 training steps. This model reached an average distance of 5.7~cm with a success rate of 49.8\% for a threshold of 5~cm, which is fairly good, but the approach presented in section~\ref{sec:sb3} will yield better results.

\begin{table}[bp]
\caption{Evaluation of the custom TD3 model. Threshold is the minimum proximity of the robot end effector to the target in order to consider an episode successful.}
\begin{center}
\begin{tabular}{|c|c|c|}
\hline
\textbf{Threshold (m)}  &\textbf{Success Rate (--)}  &\textbf{Average Distance (m)} \\
\hline
0.20 & 0.994 & 0.057 \\
0.15 & 0.979 & 0.057 \\
0.10 & 0.886 & 0.058 \\
0.05 & 0.498 & 0.057 \\ 
\hline
\end{tabular}
\label{tab:td3_custom}
\end{center}
\end{table}

\subsubsection{Stable Baselines supported implementations} \label{sec:sb3}
To obtain more comparable experiment results, we also applied more reliable, and optimized algorithm implementations from the stable-baselines3 (SB3) \cite{stable-baselines3} library, besides the custom implementations presented in section~\ref{sec:custom}.

To easily integrate these algorithms, our setup had to meet the following specifications. First of all, SB3 requires an environment that includes the agent with its functionalities and follows the Gym interface to maintain compatibility \cite{customenv_sb32020}. This interface essentially comprises five methods. The constructor of the environment class, has to specify the dimensions and value range of the action and observation space. Furthermore, one has to define the standardized methods \texttt{step()}, \texttt{reset()}, \texttt{compute\_reward()} and optionally \texttt{render()} to satisfy the interface specifications.
After successfully implementing the environment according to the Gym interface, it can communicate with our HoLLiE robot arm model in the NRP simulation without any further adjustments and thus be used directly by the SB3 algorithms.
RL algorithms are then selected in SB3 by the common name and are configured directly using the parameter list.

Having six joints to actuate, and without any further guidance, the agent struggles to solve the reaching task. In most cases, the policy converges to reaching an average cylinder position in the middle of the table (compare results of custom TD3 model in section~\ref{sec:custom}).
The SB3 experiment setup tackles this problem by introducing curriculum learning in the form of a simplified reaching task for the agent. This strategy helps the agent identify trends for successful actions. The agent's guidance is instantiated through a constrained action space, reduced number of actuated joints, and threshold scheduling.

For the simplest configuration, the robot is reduced to three actuated joints as described in action space definition $A_1$ \eqref{eq: action_3joints}.
\begin{align}
A_1 = \{low &= [-\frac{\pi}{2}, -\frac{\pi}{2}, 0, 0, 0, 0], \nonumber \\
high &= [\frac{\pi}{2}, 0, \pi, 0, 0, 0]\}
\label{eq: action_3joints}
\end{align}
This constraint has the effect that the robot arm bends over the table in all attempts, preventing all movement in the direction next to or behind the table.
The intermediate configuration lets the robot additionally actuate the joint second closest to the end effector (fifth joint), and has an action space definition $A_2$ \eqref{eq: action_4joints}.
\begin{align}
A_2 = \{low &= [-\frac{\pi}{2}, -\frac{2\pi}{3}, 0, 0, 0, 0], \nonumber \\
high &= [\frac{\pi}{2}, 0, \frac{3\pi}{4}, 0, 0, 0]\}
\label{eq: action_4joints}
\end{align}
The last configuration actuates all six available joints and results in an action space definition $A_3$ \eqref{eq: action_6joints}.
\begin{align}
A_3 = \{low &= [-\frac{\pi}{2}, -\frac{2\pi}{3}, 0, -\frac{\pi}{2}, 0, -\pi], \nonumber \\
high &= [\frac{\pi}{2}, 0, \frac{3\pi}{4}, \frac{\pi}{2}, \frac{\pi}{2}, \pi]\}
\label{eq: action_6joints}
\end{align}

All models were trained with threshold scheduling as a curriculum learning approach. In this process, all models started with an initial threshold of 20~cm. The threshold was then reduced by 2~cm in the range between 20~cm and 10~cm and by 1~cm below 10~cm whenever achieving 15 consecutive successful episodes.

After finalizing the SB3-based experiment setup, we conducted experiments using the model-free RL algorithms: TD3 and SAC. Initially, for both algorithms, we created models for the combinations of dense or sparse reward functions (Tab.~\ref{tab:reward_functions}) and with or without HER. These models actuate only the first three joints with the action space in \eqref{eq: action_3joints}. We compare the performance of these four models for TD3 and SAC, respectively, to find out which combination of reward type and HER usage leads to the best results. Interestingly, in the case of TD3, the use of HER leads to better results in both dense and sparse reward settings. However, for SAC, the comparison shows the complete opposite: the performance of the agent using HER is worse in both cases (see Tab.~\ref{tab:sb3_groundtruth} with actuated joints 1--3). 

\begin{table*}[!tp]
\caption{Evaluation of the Reinforcement Learning algorithms based on Stable-Baselines3. The threshold is the minimum proximity of the robot end effector to the target in order to consider an episode successful. The actuated joints show which joints could be used by the agent to solve the reaching task.}
\begin{center}
\begin{tabular}{|c *{11}{|c}|}
\hline
&\textbf{Actuated}  &\multirow{2}{*}{\textbf{Reward Type}}  &\multirow{2}{*}{\textbf{HER}} &\textbf{Average}   &\multicolumn{6}{|c|}{\textbf{Success Rate (--)}} &\multirow{2}{*}{\textbf{Training length $^{\mathrm{b}}$}} \\
\cline{6-11} 
&\textbf{Joints} &&&\textbf{Distance (m)} &\textbf{\textit{20~cm$^{\mathrm{a}}$}} &\textbf{\textit{15~cm$^{\mathrm{a}}$}} &\textbf{\textit{10~cm$^{\mathrm{a}}$}} &\textbf{\textit{7~cm$^{\mathrm{a}}$}} &\textbf{\textit{5~cm$^{\mathrm{a}}$}} &\textbf{\textit{3~cm$^{\mathrm{a}}$}} \\
\hline
\multirow{8}{*}{\rotatebox[origin=c]{90}{TD3}}
&1--3    &dense  &no  &0.057 &0.999 &0.978 &0.872 &N/A &0.542 &N/A &10000 \\
&1--3    &dense  &yes &0.052 &0.998 &0.985 &0.890 &N/A &0.658 &N/A &10000 \\
&1--3    &sparse &no  &0.074 &0.981 &0.907 &0.799 &N/A &0.381 &N/A &10000 \\
&1--3    &sparse &yes &0.062 &1.000 &0.975 &0.849 &N/A &0.472 &N/A &10000 \\
&\textbf{1--3, 5} &\textbf{dense}  &\textbf{yes} &\textbf{0.024} &N/A &N/A &\textbf{0.982} &\textbf{0.958} &\textbf{0.924} &\textbf{0.766} &\textbf{30000} \\
&1--6    &dense  &yes &0.046 &N/A &N/A &0.866 &0.806 &0.744 &0.534 &60000 \\
\cdashline{2-12}
&1--3, 5$^{\mathrm{c}}$ &dense  &yes &0.061 &N/A &N/A &0.940 &0.790 &0.540 &0.312 &30000 \\
&1--3, 5$^{\mathrm{d}}$ &dense  &yes &0.056 &N/A &N/A &0.918 &0.800 &0.630 &0.320 &30000 \\
\hline
\multirow{10}{*}{\rotatebox[origin=c]{90}{SAC}} 
&1--3 &dense  &no  &0.070 &0.986 &0.936 &0.758 &N/A &0.454 &N/A &5000 \\
&1--3 &dense  &yes &0.091 &0.988 &0.884 &0.568 &N/A &0.252 &N/A &5000\\
&1--3 &sparse &no  &0.054 &0.988 &0.940 &0.868 &N/A &0.646 &N/A &5000\\
&1--3 &sparse &yes &0.109 &0.866 &0.788 &0.612 &N/A &0.342 &N/A &5000\\
&1--3, 5 &dense &no &0.047 &1.000 &0.984 &0.948 &N/A&0.760  &N/A &4000\\
&1--3, 5 &dense &yes &0.086 &1.000 &0.992 &0.840 &N/A &0.688  &N/A &4500\\
&1--3, 5 &sparse &no &0.029 &N/A &N/A &0.966 &0.932 &0.866 &0.600 &4000\\
&1--3, 5 &sparse &yes &0.042 &N/A &N/A &0.924 &0.864 &0.742 &0.376 &4500\\
&\textbf{1--6} &\textbf{sparse} &\textbf{yes} &\textbf{0.021} &N/A &N/A &\textbf{0.998} &\textbf{0.978} &\textbf{0.932 }&\textbf{0.816} &\textbf{18000}\\
\hline
\multicolumn{10}{l}{$^{\mathrm{a}}$Evaluation threshold.} \\
\multicolumn{10}{l}{$^{\mathrm{b}}$The length of the training process is given in the number of episodes.} \\
\multicolumn{10}{l}{$^{\mathrm{c}}$Learning from images: manual extraction} \\
\multicolumn{10}{l}{$^{\mathrm{d}}$Learning from images: CNN extraction}
\end{tabular}
\label{tab:sb3_groundtruth}
\end{center}
\end{table*}

The TD3 models using only three joints converged after around 10000 episodes. The comparison shows that the best performing configuration is training with dense rewards and using HER. This configuration is the same as in the case of the custom TD3 implementation in section~\ref{sec:custom}. The average distance is 5.2~cm with a success rate of 65.8\% for a threshold of 5~cm. This result is slightly better than that of the custom TD3 implementation (Tab.~\ref{tab:td3_custom}), which demonstrates that the SB3 algorithms are more reliable and optimized.

Consequently, for TD3, we use the dense reward function and HER. In the next step, a TD3 model actuating four joints with the action space $A_2$ \eqref{eq: action_4joints} has been trained extensively for 30000 episodes. After 10000 episodes, the improvement rate of the agent's performance was small but still significant. In the end, this model achieved an impressive performance with an average distance to target of 2.4~cm along with a success rate of 92.4\% for 5~cm threshold and 76.6\% for 3~cm threshold.

To train a TD3 model actuating all six joints, we relax the action space constraint of the previous model to $A_3$ \eqref{eq: action_6joints} and continue the training process. Initially, performance drops sharply as the agent must learn to deal with two additional joints. Yet again, after extensive training for 30000 episodes, the agent's results improve as in Tab.~\ref{tab:sb3_groundtruth}. Interestingly, this agent can not compete with the TD3 agent actuating four joints. This reflects the fact that the number of possible configurations increases exponentially with the dimensionality, which makes it more unlikely for the agent to find a better solution.

The three-joint SAC models were each trained for 5000 steps. The algorithms managed to learn successful policies after 2200 steps when using HER and approximately 3000 steps without using HER. Regarding the convergence time, no serious difference between sparse and dense reward models could be found.
Looking at the average distance of the models, the configuration with dense reward and no HER reaches about 7~cm and performs slightly better than dense reward with HER configuration with about 9.1~cm distance. In the experiments with sparse rewards the configuration without HER reaches an average distance of 5.4~cm and performs significantly better than the configuration with HER with about 10.9~cm.

\begin{figure*}[!bp]
\centering
\subfloat[TD3, 4 actuated joints]{\includegraphics[width=0.35\linewidth]{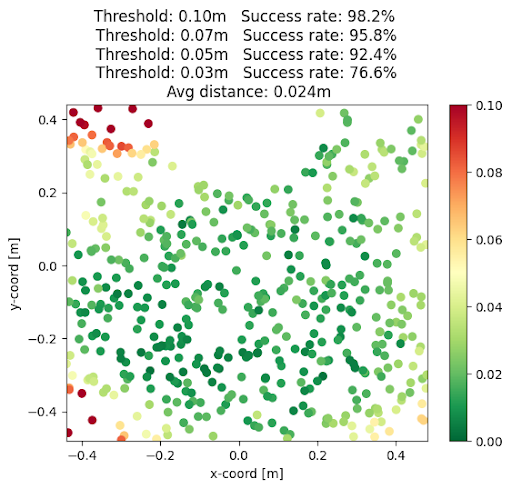}
\label{fig:td3_4joints}}
\hfil
\subfloat[SAC, 6 actuated joints]{\includegraphics[width=0.35\linewidth]{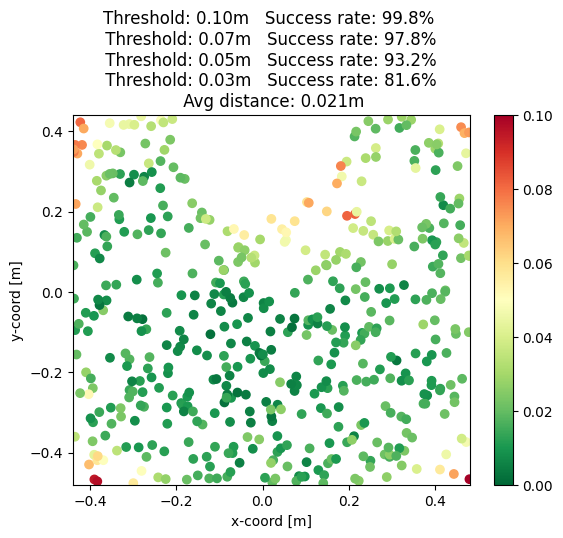}
\label{fig:sac_4joints}}
\caption{Best performing TD3 and SAC models. The points in the scatter plots are the target positions, and they are colored based on the achieved proximity of the end effector. The empty semicircle on the top of the plots is where the robot's base is mounted to the table. A target position in this area would result in the collision of the robot with itself. The TD3 model was trained with dense rewards and HER, while the SAC model was trained with sparse rewards.}
\label{fig:sb3_scatters}
\end{figure*}
The four-joint models were trained for 4000 steps (without HER) and 4500 steps (with HER). Even in the configuration with four actuated joints, the models converge before the end of the scheduled training time. Looking again at the performance of the models, the configuration with dense reward and no HER  
evaluates with an average distance of 4.7~cm and is therefore again better as the configuration with use of HER with an average distance of 8.6~cm. The experiments with sparse reward reach average distances of 2.9~cm in the setting without HER and around 4.2~cm in the one with HER and therefore approximately half of the distance as in the dense setting.

The six-joint model was trained for 18000 steps. In contrast to the six-joint TD3 model, this model was trained from scratch, but also uses the constrained action space described in $A_3$ \eqref{eq: action_6joints}. Despite the relatively short training time the model reached an average distance of 2.1~cm and performs a little bit better than the SAC and TD3 four-joint models.

An overview of all achieved results is shown in Tab.~\ref{tab:sb3_groundtruth}. Fig.~\ref{fig:sb3_scatters} shows the best performing TD3 and SAC models in action.

\subsection{Learning from images}
There are generally three approaches when learning from images:
\begin{enumerate}
  \item Extract the ground truth data from images and feed it to our pre-trained agents.
  \item Extract a latent space representation with decreased dimensionality from the images and feed it as input to the agents.
  \item Directly feed images as input to the agents.
\end{enumerate}
While learning from ground truth data reduced the agent's task to finding an approximation of the inverse kinematics, learning from images or latent space representations of these images is significantly more complex.
On the one hand, there is the increased computational cost as the raw image and even the latent space representation will be considerably larger than the two or three values cylinder coordinate representation. On the other hand, and more crucial, the agent now has to find an additional transformation from image or latent space to a cartesian representation of the cylinder location before learning an approximation of the inverse kinematics.
To simplify the setting we use a top view camera that looks down at the table from above. Since the cylinder always spawns on the table surface, this setup reduces the extraction of the cylinder location from 3D to 2D.

\subsubsection{Extraction of the ground truth}
One intuitive approach to learn from images is the extraction of ground truth data from the image. In this case, we can reutilize the pre-trained models on ground truth. To achieve this, we followed two general approaches. First, we use classic computer vision (CV) to extract the ground truth. Then, we also use a deep learning (DL) method, training a convolutional neural network (CNN) to recognize the location of the cylinder. 

\begin{figure}[tp]
\centering
\subfloat[Masked and Thresholded]{\includegraphics[width=0.235\textwidth]{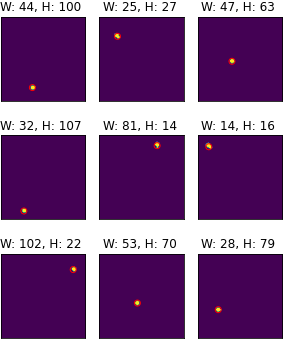}
\label{fig:manual_blob}}
\hfil
\subfloat[Projected to Image]{\includegraphics[width=0.235\textwidth]{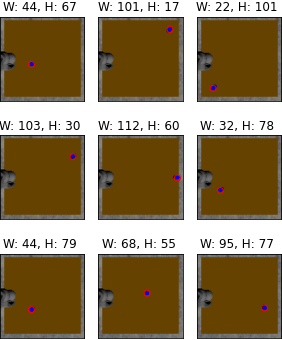}
\label{fig:manual}}
\caption{Manual extraction of the cylinder location. (a) Masked and thresholded images. A red circle marks the remaining white blob. (b) Projection of the red circle into the original images. The images in (a) and (b) do not correspond.}
\label{fig:manual_extraction}
\end{figure}

To manually extract the ground truth, we first created a mask by averaging over a set of images that do not contain the cylinder. Afterwards, we subtracted the mask from each query image and applied a threshold that left us with a blob representing the cylinder (Fig.~\ref{fig:manual_blob}). After extracting the location of this blob from the matrix, we transformed the coordinates from the image to the simulation coordinate frame.

\begin{figure*}[!bp]
\centering
\subfloat[TD3, image input, manual extraction]{\includegraphics[width=0.35\linewidth]{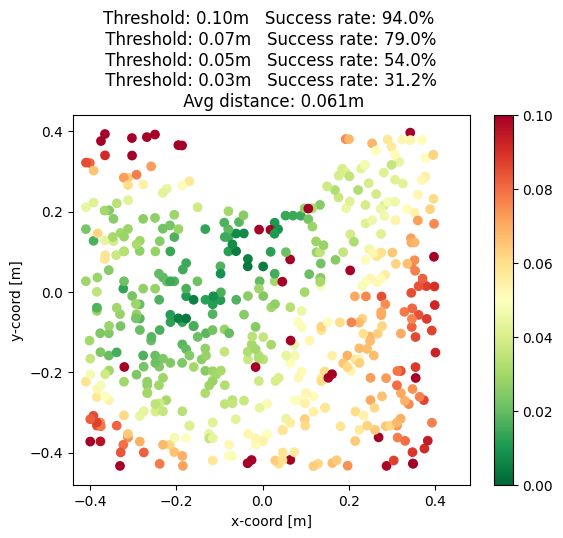}
\label{fig:td3_img_4joints}}
\hfil
\subfloat[TD3, image input, CNN extraction]{\includegraphics[width=0.35\linewidth]{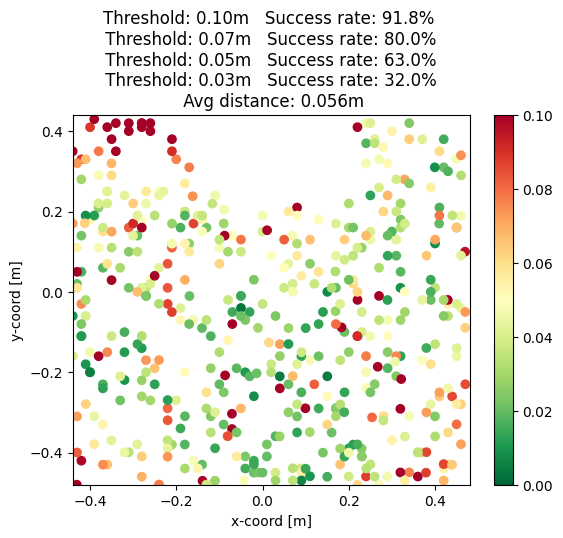}
\label{fig:td3_cnn_4joints}}
\caption{Performance of a TD3 agent, with four actuated joints, on the \textbf{image data} processed (a) by the manual extraction approach and (b) by the CNN extraction approach. The points in the scatter plots are the target positions - their color correlates with the achieved proximity of the end effector. The empty semicircle on the top of the plots is where the robot's base is mounted to the table. A target position in this area would result in the collision of the robot with itself. The TD3 model was trained with dense rewards and HER.}
\label{fig:image_scatters}
\end{figure*}

The calculation of the average error for 3000 samples returned a deviation of 20~mm in the x-direction and 9~mm in the y-direction. This error is primarily caused by the transformation from the image to the simulation frame and the choice of the center of the cylinder blob. However, while the coordinate extraction using the manual method generally differed from ground truth by less than 2~cm, the effective error margin turned out to be $\pm$2~cm. Fig.~\ref{fig:td3_img_4joints} shows the performance of our best performing TD3 model (trained on ground truth) using the manual extraction pipeline.

For the second approach, we trained a convolutional neural network (CNN) to extract the location of the cylinder. Because of the nearly invariant data input (only the cylinder location changes in the entire image) it is hard to train the CNN to recognize any geometrical features (e.g. circles). Therefore, we first pre-trained the convolutional layers of the CNN using the CIFAR-10 dataset \cite{CIFAR-10}. Afterwards, we used our self-generated dataset of 2000 images to fine-tune the network.

\begin{figure}[H]
    \centering
    \includegraphics[width=\linewidth]{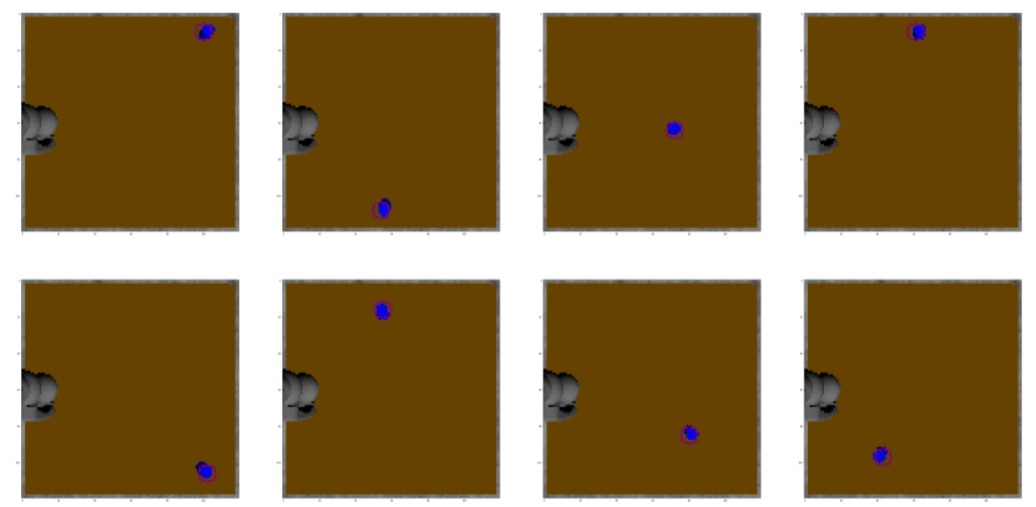}
    \caption{Cylinder position prediction with CNN. The actual cylinders are presented in the picture as the blue cylinders. The predicted positions of the cylinder are presented as the purple circles with the centers as the values of the prediction and a radius of 5 pixels. }
    \label{fig:basic_cnn}
\end{figure}

\noindent The relatively simple network converges quickly and shows an average error of 13~mm in the x-direction and 25~mm in the y-direction for 250 validation dates. Fig.~\ref{fig:basic_cnn} shows the accuracy of the CNN predictions. We evaluate the CNN model with our best performing TD3 four-joint model (Fig.~\ref{fig:td3_cnn_4joints}). The results prove the CNN is highly sensitive to geometrical error. For example, if the cylinder spawns on its side or the robot arm is not in its initial position in the image, the CNN is not likely to be able to calculate an accurate output. The failing instances with high achieved reaching proximity in the results are caused by this type of error. 

\subsubsection{Latent space extraction}

The extraction of the ground truth is relatively rigid. The handcrafted extraction of the coordinates relies on little to no variation in the images to be able to mask everything but the cylinder. Additionally, the chosen thresholds are intensity-dependent. Accordingly, adopting this approach to new settings requires quite a bit of manual work or may not even be possible. The DL-based pipeline is already less rigid as one can simply unfreeze the fully connected layers and retrain the model to adapt to the new setting. However, the DL pipeline also relies on little variation in the images to accurately extract the cylinder location once it learned to identify such an object. Moreover, since the DL-based approach is supervised it relies on the actual position of the cylinder to be known.  
The latent space extraction solves at least the latter problem as it follows an unsupervised paradigm. Instead of extracting the cylinder location, we try to drastically reduce the dimensionality of the image and find a latent representation that encodes the most meaningful features of the image. The hope is that the position of the cylinder gets encoded into this reduced space representation. 

Theoretically, this approach has the benefit that we can apply it to various settings and tasks. Although, it will require considerable retraining and tuning.
While the little variation in the image data was crucial for the direct extraction of the ground truth, it became the pitfall for an autoencoder-based latent space extraction. Instead of encoding the cylinder location, the autoencoder effectively became a denoiser and filtered out the cylinder that made up less than five percent of the whole image.
% (Fig.~\ref{fig:auto_encoder}). 

% \begin{figure}[bp]
% \centering
% \includegraphics[width=0.45\textwidth]{figures/auto_filter.png}
% \label{fig:auto_filter}
% \caption{The upper row are the original images. The lower row shows the reconstructed images. As can be seen, the cylinder is filtered out.}
% \label{fig:auto_encoder}
% \end{figure}

Extensive tuning and image preprocessing could not overcome the autoencoder's denoising effect. However, compared to the CNN-based approach, the task of the autoencoder is also considerably less straighforward. On the one hand, the CNN has a two-valued output and trains directly to extract the 2D coordinates. On the other hand, the autoencoder learns to reconstruct the input image from a latent spatial representation, which ideally encodes the cylinder coordinates.

\subsubsection{Images as direct input}

Stable Baselines3 also supports a CNN policy for the aforementioned RL algorithms, whose inputs are images. Multiple inputs can be used when the observation space contains both images and joint angle values. This CNN policy was tested with the SAC algorithm. Although the robot arm was able to move slightly in the direction of the cylinder, the final performance is not yet satisfactory and still subject to development.

\section{Discussion and Conclusion}

Considering the achieved results, it can be summarized that both models, either trained with ground truth data or with images, have achieved notable results in solving the reaching task.
Using the default parametrization of the SB3 algorithms, SAC converges constantly faster than TD3, as can be seen by comparing the training times in Tab.~\ref{tab:sb3_groundtruth}. 

Contrary to our expectations, the three-joint SAC models using HER did not achieve superior performance. One possible reason could be that the chosen action space is quite constrained, and therefore, training without HER is anyway not suffering from bad sample-efficiency. In contrast to this, TD3 performed better when using HER.
Additionally, the SAC configurations using sparse rewards worked better than those with dense rewards, contrary to TD3 models.
The results of the six-joint SAC model showed that HER can lead to impressive results in combination with SAC as well.

Further, the limitation of the action space, the number of accessible joints, and the threshold scheduling had a significant impact on the training time. Interestingly, the four-joint TD3 model could cross small thresholds with much greater confidence and less training time than the six-joint TD3 model. In general, however, we must assume that a more extensive grid search and longer learning cycles would result in a predominant six-joint setting. 

When learning from images, it proved very effective to use a top-mounted camera to reduce the given problem of the localization of the cylinder from 3D to 2D. However, while the coordinate extraction using the DL-based approach and the manual method generally differed from ground truth by less than 2~cm, the overall performance drop for the reaching task was around 4~cm. A more precise coordinate conversion and finer discretization could therefore have a substantial impact on the error margin. 
Regarding the extraction of a latent space representation, we must conclude that a classical autoencoder may not be the right tool for images such as ours. Intensive tuning and various tricks could not overcome the denoising effect of our autoencoder. A possible solution could be changing the architecture to a variational autoencoder or use a pre-trained model on a dataset with large image variation. 

\section*{Acknowledgment}

First and foremost, we want to thank our tutors Florian Walter, Mahmoud Akl, and Josip Josifovski. We would like to express our sincere gratitude to you for sharing your expertise and advice. Further, we want to thank TUM for providing us with considerable computational resources via the Leibniz-Rechenzentrum.

% See: https://tex.stackexchange.com/questions/442087/format-a-bib-in-ieee-style-automatically
\bibliographystyle{IEEEtran}
\bibliography{IEEEabrv,mybib}

\end{document}